%% file: 2025_ICIT.tex
\documentclass[conference, hidelinks]{IEEEtran}
\IEEEoverridecommandlockouts

\usepackage{cite}
\usepackage{amsmath,amssymb,amsfonts}
\usepackage{algorithmic}
\usepackage{graphicx}
\usepackage{textcomp}
\usepackage{xcolor}

\usepackage{orcidlink}
\usepackage{flafter} 
\usepackage{placeins} 
\usepackage{hyperref} 
\usepackage{subcaption}

\definecolor{UStuttDarkBlue}{RGB}{0,81,158}
\definecolor{UStuttDarkOrange}{RGB}{228,175,52}

\usepackage{tikz}
\usetikzlibrary{calc,fit, positioning,arrows.meta}
\tikzset{>={Latex[width=2mm,length=2mm]}} 
\tikzstyle{block} = [draw=black, fill=white, rectangle, align=center, minimum height=1em, minimum width=2em, font = {\footnotesize}]
\tikzstyle{sum} = [draw, circle, node distance=1cm]
\tikzstyle{input} = [coordinate]
\tikzstyle{output} = [coordinate]
\tikzstyle{punkt} = [circle,fill,inner sep=0pt,minimum size=5pt]
\tikzstyle{gray dotted} = [draw=gray, fill=white, rectangle, align=center, line width=1pt, dash pattern=on 1pt off 4pt on 6pt off 4pt, inner sep=3mm, rounded corners, font = {\small}]
\tikzstyle{blueblock} = [draw=UStuttDarkBlue, fill=white, rectangle, align=center, minimum height=1em, minimum width=2em, line width =1.5pt, font = {\footnotesize}]
\tikzstyle{orangeblock} = [draw=UStuttDarkOrange, fill=white, rectangle, align=center, minimum height=1em, minimum width=2em, line width =2pt, font = {\footnotesize}]

\usepackage{siunitx} 
\DeclareSIUnit{\Bit}{Bit}

\def\BibTeX{{\rm B\kern-.05em{\sc i\kern-.025em b}\kern-.08em
    T\kern-.1667em\lower.7ex\hbox{E}\kern-.125emX}}

\begin{document}

\title{Modeling Load-, Velocity-, and Temperature-\\Dependent Transmission Errors of Cycloidal Drives for Industrial Robots Using Fourier Series
	\thanks{Funded by the Deutsche Forschungsgemeinschaft (DFG, German Research Foundation) – 443677015\\
	Data set available online:
	\url{https://doi.org/10.18419/DARUS-4454}
	}
}

\author{
	\IEEEauthorblockN{Christian J.\,E. Bauer\IEEEauthorrefmark{1} \orcidlink{0009-0002-7979-3877},
		Valentin Kamm\IEEEauthorrefmark{1}
		\orcidlink{0000-0003-4748-5415},
		Marcel Dzubba\IEEEauthorrefmark{1} \orcidlink{0009-0008-2054-6365},\\
		Lukas Steinle\IEEEauthorrefmark{1} \orcidlink{0000-0002-6035-0067},
		Armin Lechler\IEEEauthorrefmark{1} \orcidlink{0000-0002-4073-1487},
		Alexander Verl\IEEEauthorrefmark{1} \orcidlink{0000-0002-2548-6620}}
	\IEEEauthorblockA{\IEEEauthorrefmark{1}\textit{Institute for Control Engineering of Machine Tools and Manufacturing Units~(ISW)},\\
		University of Stuttgart, 70174 Stuttgart, Germany\\
		E-mail: christian.bauer@isw.uni-stuttgart.de}
	}
\maketitle

\begin{abstract}
Industrial robots are rarely used for machining tasks due to their limited path accuracy. This accuracy is mainly limited by inaccuracies in the drive trains. Compliance and transmission errors occur in the joint gearboxes. While transmission errors have been extensively studied for strain wave gears, there is little research on these errors in cycloidal drives. This gearbox type is commonly used in industrial robots for medium to heavy payloads. It is proposed to model the mainly periodic transmission errors using a Fourier series where amplitude and phase are defined as a polynomial function of the main influence factors load-torque, velocity, and temperature. Measurements of the transmission errors were conducted using an experimental setup representing a single robot joint. In the evaluation of the measurement data, harmonic frequencies were related to mechanical properties of the cycloidal drive. These frequencies were used to identify the parameters of the polynomial Fourier series model. Compared to validation measurements, the derived model shows an average root mean square error of 0.026 mrad. It is proposed to use the output of the resulting model in a feedforward control approach to compensate the transmission errors and to increase the path accuracy of industrial robots.
\end{abstract}


\section{Introduction}
The application of industrial robots~(IR) for machining tasks can provide advantages over the use of classical machine tools due to their high flexibility, high degree of freedom, and a favorable workspace to cost ratio.
Currently, IR applications in machining are limited by an insufficient path accuracy~\cite{karimAnalysisDynamicBehavior2017}.
Compared to a standard machine tool, it is lower by a factor of~\num{100}~to~\num{1000}~\cite{zaehImprovementMachiningAccuracy2014}.
This applies to both manufacturing processes with high process forces, such as milling, and those with low process forces, such as laser cutting.
The main contributor to these path errors are the drive trains of the IR, which include permanent-magnet synchronous motors~(PMSM) and gearboxes~\cite{sweetRedefinitionRobotMotion1984, rudermanSensorlessTorsionControl2016}.
The angular position error of drive trains is influenced by compliance and kinematic transmission errors~(TEs) in the gearbox as well as torque ripple of the motor~\cite{rendersKinematicCalibrationGeometrical1991}.
The individual errors accumulate along the serial kinematics and are amplified as the axis length increases.
The use of load-side encoders to compensate for these effects leads to increased costs and system complexity.
Typically, lightweight robots contain strain wave gears~(also called harmonic drives) while IRs for heavy payloads are equipped with cycloidal drives~(CDs)~\cite{mesmerChallengesLinearizationbasedControl2021}.
Previous model-based compensation approaches for compliance effects in CDs have shown a reduction of the tracking error by~\qty{81}{\percent}~\cite{mesmerInvestigationCompensationHysteresis2023}. TEs are difficult to model and have not been considered in these previous works. This paper focuses on modeling state-dependent TEs in CDs and investigating key influence factors. In future works, a experimental closed-loop evaluation of the proposed compensation approach and a combination of the two compensation strategies are planned.

\subsection{State of the Art}
\label{ssec:StateOfTheArt}
\begin{figure}
	\centering
	\includegraphics[scale=1]{"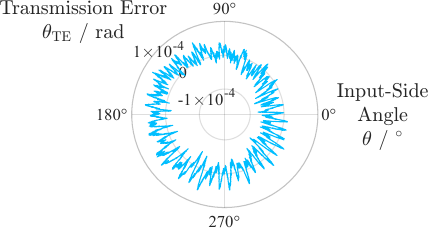"}
	\caption{Transmission errors~\(\theta_\mathrm{TE}\) without mean for \qty{1000}{\N\m} load torque, \qty{-2.5}{\degree\per\second} velocity and \qty{27}{\degreeCelsius} lubricant temperature dependent on the angle~\(\theta\).}
	\label{fig:intro_polar_plot}
\end{figure}

TEs are angle-dependent, mainly periodic errors between the motor and load-side angle of the gearboxes in the drive train caused by manufacturing and assembly tolerances~\hbox{\cite{blancheCycloidDrivesMachining1989, wuKinematicErrorAnalysis2020}}.
Their angular dependency is visualized in~\autoref{fig:intro_polar_plot} for an example measurement with the experimental setup, which will be described in~\autoref{ssec:experimental_setup}.
In literature, these errors are commonly modeled using purely angle-dependent Fourier series.
\textsc{Ghorbel} et~al. show an additional load-torque and velocity dependency for TEs in strain wave gears~\cite{ghorbelKinematicErrorHarmonic2001}.
A lubricant temperature dependency of gear errors due changes in the viscous friction is described in~\cite{qiuReviewPerformanceTesting2021}.
\textsc{Iwasaki} et~al. and \textsc{Sasaki} et~al. implement a model-based compensation for TEs~\cite{iwasakiModelingCompensationAngular2009, sasakiMETHODCOMPENSATINGFORANGULAR2011, sasakiMETHODCOMPENSATINGANGULAR2012} while \textsc{Han} et~al. use feedback from an acceleration sensor with a peak filter~\cite{hanSuppressionVibrationDue2008} and \textsc{Iwasaki} et~al. develop an adaptive notch filter for the torque setpoint to suppress vibration caused by~TEs~\cite{iwasakiVibrationSuppressionAngular2014}.
\textsc{Trung} et~al. and \textsc{Mesmer} et~al. apply load-side position sensors to suppress these vibrations~\cite{trungHInfinityControlbasedVibration2018, mesmerDriveBasedVibrationDamping2020, mesmerGainScheduledDrivebasedDamping2022, mesmerRobustDesignIndependent2022}.
While most works consider strain wave gears~\cite{iwasakiModelingCompensationAngular2009, sasakiMETHODCOMPENSATINGFORANGULAR2011, sasakiMETHODCOMPENSATINGANGULAR2012, trungHInfinityControlbasedVibration2018, hanSuppressionVibrationDue2008, iwasakiVibrationSuppressionAngular2014}, there is little research on TEs in CDs.
\textsc{Zhang} et~al. describe TEs in CDs as angle-dependent Fourier series~\cite{zhangMathematicalModelAnalysis2011} and \textsc{Kawahara} et~al., \textsc{Yoshioka} et~al., and \textsc{Hirano} et~al. present disturbance observers based on that concept~\hbox{\cite{kawaharaVibrationSuppressionFeedback2013, yoshiokaVibrationSuppressingControl2014, hiranoVibrationSuppressionControl2016}}.
\textsc{Xu} et~al. and \textsc{Wang} et~al. develop analytical models of the CD geometry to predict the TE~\hbox{\cite{xuDesignDynamicTransmission2023, wangTransmissionCharacteristicsRV2024}}.
\textsc{Wang} et~al. propose an analytical model that combines TEs and lost motion~\cite{wangPositioningAccuracyPrediction2024}.
The velocity and load dependency of TEs can also be shown for rack-and-pinion drives~\cite{steinleExperimentalInvestigationImplications2022}. In~\cite{verlAdaptiveCompensationTransmission2022}, \textsc{Verl} and \textsc{Steinle} realize a compensation based on regression trees while in~\cite{steinleLearningCompensationStateDependent2024}, a combination of neural networks and regression trees is implemented.
\textsc{Bilancia} et~al. analyze the velocity and temperature dependency of TEs in CDs and propose to vary the amplitude and phase of the Fourier series as a function of these influences~\cite{bilanciaAccurateTransmissionPerformance2022}.
In~\cite{bilanciaOnlineMotionAccuracy2025}, the influence of the load torque is additionally modeled by \textsc{Bilancia} et~al. using various machine learning models to estimate amplitude and phase. These compensation approaches use the load-side angle as an input, which cannot be measured in a standard robot.

\subsection{Contribution}
In this work, a TE modeling approach for CDs based on input-side angle and velocity from the motor encoder is presented. At this point, the proposed model still relies on the load-side torque and the lubricant temperatures. In this contribution, these properties are provided by sensors, while in future works an observer-based implementation is planned. In the planned concept, the compensation parameters are identified in an end-of-line measurement using a load-side encoder, a load-side torque sensor and a lubricant temperature sensor, while in operation only the robot's input-side encoders are required. In addition, in the present work, the relationship between mechanical properties of CDs and harmonic frequencies of TEs are analyzed in detail. A Fourier series model with a polynomial formulation of amplitude and phase as a function of load torque, velocity, and temperature is proposed.

\section{Experimental Setup and Methods}
\begin{figure}
	\centering
	\includegraphics[scale=1.0]{"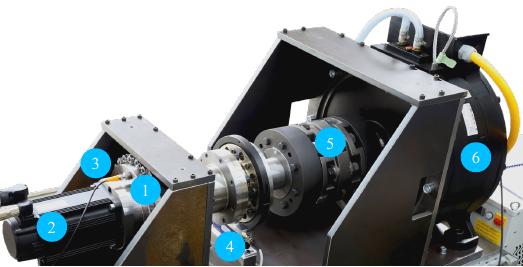"}
	\caption{Experimental setup with 1)~cycloidal drive, 2)~input-side motor, 3)~temperature sensor, 4)~torque sensor, 5)~coupling, and 6)~load-side motor~(presented in~\cite{mesmerModelingIdentificationHysteresis2022}).}
	\label{fig:experiment-setup}
\end{figure}

To analyze effects of TEs on an isolated robot joint, experiments were carried out with the setup shown in~\autoref{fig:experiment-setup}.
Next, a polynomial Fourier series formulation for the TEs was developed and its parameters were determined in a subsequent step.

\subsection{Isolated Robot Joint for Experiments}
\label{ssec:experimental_setup}
In the experimental setup a Nabtesco \hbox{\textit{RH380-N}} CD with a rated output torque of~\qty{3724}{\N\m} and the parameters listed in~\autoref{tab:CD-param} is investigated.
It combines a planetary gear stage and a CD stage.
In the planetary gear stage, an input gear with~\hbox{\(z_{\num{1}}=\num{24}\)} teeth drives three spur gears with~\(z_{\num{2}}=\num{96}\) resulting in a reduction ratio of~\(u_{\num{1}}=\frac{z_{\num{2}}}{z_{\num{1}}}=\num{4}\) for the first stage.
The second CD stage compromises a cycloidal disc with~\(z_{\num{3}}=\num{45}\) lobes and~\(z_{\num{4}}=\num{46}\) pins, resulting in a reduction ratio of~\(u_{\num{2}}=\num{46}\).
Including the additional rotation of the output shaft the total reduction ratio becomes~\(u=\num{1}+\frac{z_{\num{2}}}{z_{\num{1}}} \cdot z_{\num{4}} = \num{185}\)~\cite{RVNSeriesProduct}.
\begin{table}
	\caption{Gearbox Components and Factors of a Nabtesco \hbox{\textit{RH380-N}} Cycloidal Drive~\cite{RVNSeriesProduct}}
	\begin{center}
		\begin{tabular}{|c|c|c|}
			\hline
			\textbf{Component} & \textbf{Symbol} & \textbf{Factor} \\
			\hline
			Teeth of input gear & \(z_{\num{1}}\) & \num{24} \\
			Teeth of spur gears & \(z_{\num{2}}\) & \num{96} \\
			Lobes of cycloidal disc & \(z_{\num{3}}\) & \num{45} \\
			Pins of cycloidal disc & \(z_{\num{4}}\) & \num{46} \\
			\hline
			Planetary gear stage reduction ratio & \(u_{\num{1}}\) & \num{4} \\
			Cycloidal gear stage reduction ratio & \(u_{\num{2}}\) & \num{46} \\
			Total reduction ratio & \(u\) & \num{185} \\
			\hline
		\end{tabular}
		\label{tab:CD-param}
	\end{center}
\end{table}
The gearbox is driven by a PMSM~(\textit{MSK070D}, Bosch Rexroth~AG) with \qty{17.5}{\N\m} nominal and \qty{52.5}{\N\m} maximum torque.
The position of this input-side motor is measured by an encoder with a \qty{13}{Bit} resolution.
The temperature of the gearbox lubricant is measured with a PT100 sensor.
The analog torque sensor~(\textit{HBM~T40B}, Hottinger Brüel~\& Kjaer GmbH), which measures the output-side torque of the gearbox, has a rated torque of~\qty{10}{\N\m}.
Its signal is digitized with \qty{16}{Bit} resolution.
A coupling~(\textit{ROBA-DS~1400}, Chr. Mayr~GmbH~+~Co.~KG) with \qty{15}{\kilo\N\m\per\radian} torsional stiffness is applied to connect the load-side motor~(\textit{DST2-315KO}, Baumüller Nürnberg~GmbH).
This high-torque motor with~\num{25}~pole pairs has a rated torque of~\qty{1650}{\N\m} and can exert~\qty{3400}{\N\m} at its maximum.
Its encoder~(\textit{ECN~1325}, DR. JOHANNES HEIDENHAIN~GmbH, \qty{25}{Bit} resolution) is used to derive the load-side position of the system.
The systems control architecture is implemented on a Speedgoat rapid control prototyping system.
It is connected to the motor drives via an EtherCAT bus with a cycle time of~\qty{2}{\kilo\Hz}~\cite{mesmerModelingIdentificationHysteresis2022}.

\subsection{Modeling of the Transmission Errors}
In this contribution TEs are denoted by
\begin{align}
	\theta_\mathrm{TE} &= q - \theta\mathrm{,}
\end{align}
where the input-side angle~\(\theta\) is divided by~\(u\) to fit the scaling of the load-side angle~\(q\).
It is assumed that TEs are periodic~\cite{ghorbelKinematicErrorHarmonic2001} and can be modeled by a Fourier series in the amplitude-phase form
\begin{align}
	\hat{\theta}_\mathrm{TE} &=
	\sum_k {A_k \cdot
		\sin \left( 2\pi \cdot f_k \cdot \frac{\theta}{\theta_\mathrm{period}} + \varphi_k\right) } + a_0\mathrm{,}
\end{align}
considering~\(k\) harmonic frequencies~\(f_k\) at which the series has amplitude~\(A_k\), phase~\(\varphi_k\), and constant offset~\(a_0\).
The spatial frequencies~\(f_k\) are defined as dimensionless fractions of a full load-side revolution~(period~\(\theta_\mathrm{period} = 2\pi\)). These~\(f_k\) describe how often the component~\(k\) of the TEs occurs per load-side revolution.
This definition simplifies the Fourier series to
\begin{align}
	\hat{\theta}_\mathrm{TE} &=
	\sum_k {A_k \cdot
		\sin \left(f_k \cdot \theta + \varphi_k\right) } + a_0\mathrm{.}
\end{align}
The model assumption that TEs depend on load-side torque~\(\tau_\mathrm{load}\), input-side velocity~\(\dot{\theta}\), and gearbox lubricant temperature~\(T\) was derived from the literature~(\autoref{ssec:StateOfTheArt}). Therefore, the properties~\(A_k\),~\(\varphi_k\), and~\(a_0\) are also state-dependent. A quadratic polynomial of the form
\begin{align}
	A_k \left(\tau_\mathrm{load}, \dot{\theta}, T\right)
	&= p_{A,k,1} \cdot \tau_\mathrm{load}^2 + p_{A,k,2} \cdot \dot{\theta}^2 + p_{A,k,3} \cdot T^2 \nonumber \\
	&+ p_{A,k,4} \cdot \tau_\mathrm{load} + p_{A,k,5} \cdot \dot{\theta} + p_{A,k,6} \cdot T \nonumber \\
	&+ p_{A,k,7} \cdot \tau_\mathrm{load} \cdot \dot{\theta} + p_{A,k,8} \cdot \tau_\mathrm{load} \cdot T \nonumber \\
	&+ p_{A,k,9} \cdot \dot{\theta} \cdot T + p_{A,k,10}\mathrm{,}
\end{align}
with the parameters \(p_{A,k,i}\) and \(i\in\{1,\dots,10\}\), is introduced to describe these influences for~\(A_k\), taking into account two-factor interactions.
A second order polynomial is chosen as a compromise between coverage of nonlinear behavior and number of parameters. 
Equivalent polynomials are used for~\(\varphi_k (\tau_\mathrm{load}, \dot{\theta}, T)\) and~\(a_0(\tau_\mathrm{load}, \dot{\theta}, T)\) with \(p_{\cdot,k,i}\) and \(i\in\{1,\dots,10\}\).

\subsection{Parameter Identification}
\label{ssec:identification}
To identify the parameters of the system model, a constant low velocity was commanded to the input-side motor while a constant load was applied with the load-side motor.
The temperature of the gearbox lubricant was increased with heating sequences in between the measurements, where high velocities were commanded to the input-side motor in order to generate heat within the gearbox.
This method is beneficial compared to an external heating, as the gearbox would also heat up from the inside during industrial operation.
In the first step of the parameter identification process,~\(A_k\),~\(\varphi_k\), and~\(a_0\) for each measurement were calculated.
In the second step, the parameters of the polynomial were optimized to fit the single measurements.
In this process, two separate sets of parameters were calculated for forward and backward motion because the TEs differ depending on the direction of motion.\\
The~\(A_k\) were derived from the single-sided amplitude spectrum in the spatial frequency domain.
To calculate the spectrum in spatial frequency domain, the measurement points must be equidistantly spaced along the~\(\theta\)~axis.
Although a constant velocity is commanded, a perfectly uniform distribution is not guaranteed.
Therefore, the data was linearly interpolated first.
Linear trends were removed and a Hamming window was applied to avoid leakage.
In a further preprocessing step, the measurement data was zero-padded to the next higher power of~\num{2} above \num{10} times its original length.
This increases the frequency resolution of the fast Fourier transform~(FFT), which is equal to an ideal interpolation with the \(\mathrm{sinc}\) function~\cite{smithMathematicsDiscreteFourier2007}.
The two-sided amplitude spectrum was calculated from the FFT and divided by a window normalization factor before it was converted into the single-sided spectrum.
The~\(\varphi_k\) were determined from the equivalent sine-cosine form of the Fourier series in the spatial domain
\begin{align}
	\hat{\theta}_\mathrm{TE} =\sum_k \Biggl( a_k \cdot \cos \left(f_k \cdot \theta \right) + b_k \cdot \sin \left(f_k \cdot \theta  \right)
	\Biggr) + a_0\mathrm{,}
\end{align}
with the parameters
\begin{align}
	a_k &= \frac{1}{\pi}
	\cdot \int_{0}^{2\pi} \theta_\mathrm{TE} (\overline{\theta}) \cdot \cos \left(f_k \cdot \theta_\mathrm{TE} (\overline{\theta}) \right) \mathrm{d}\overline{\theta}
\end{align}
and
\begin{align}
	b_k &= \frac{1}{\pi}
	\cdot \int_{0}^{2\pi} \theta_\mathrm{TE} (\overline{\theta}) \cdot \sin \left(f_k \cdot \theta_\mathrm{TE} (\overline{\theta}) \right) \mathrm{d}\overline{\theta}\mathrm{.}
\end{align}
These parameters~\(a_k\) and~\(b_k\), which were directly calculated from the measurements of~\(\theta_\mathrm{TE}\), can be converted into the phase using the 2-argument arcus tangent
\begin{align}
	\varphi_k &= \mathrm{atan2}\left(a_k,b_k \right)\mathrm{.}
\end{align}
The constant offset~\(a_0\) was calculated as mean value of the TEs~\(\theta_\mathrm{TE}\).\\
In a preliminary step before estimating the polynomial parameters, the input values~\(\tau_\mathrm{load}\),~\(\dot{\theta}\) and~\(T\) were standardized by removing the mean and dividing by the standard deviation in order to avoid an imbalance due to the different scaling.
Then, for each of the harmonic frequencies~\(f_k\), the polynomial parameters~\(p_{\cdot,k,i}\) for~\(i=\{1,\dots,10\}\) were obtained using the linear least squares method.
The resulting parameters were scaled accordingly so that their scaling fits the raw input data rather than the standardized data.\\
Based on the analysis of measurement data, relevant harmonic frequencies were selected for the TE Fourier series model.
The harmonic frequencies with high amplitudes can be related to the mechanical properties of the CD presented in~\autoref{ssec:experimental_setup}.
Since an overcompensation of the TEs or an incorrect phase of the TE model would result in worse system performance, it is important to ensure that the output of the compensation is not larger than the actual TEs.
Therefore, only reliable information should be included in the model, so only harmonic frequencies that are related to a mechanical parameter were considered.

\section{Results}
In this section, the properties of TEs in CDs and their relationship to mechanical properties of the CD are analyzed based on collected measurement data. Afterwards, the output of the proposed polynomial Fourier series model is compared with independent validation data.

\subsection{Transmission Errors}
\begin{figure*}
	\centering
	\begin{subfigure}{0.495\textwidth}
		\includegraphics[scale=1]{"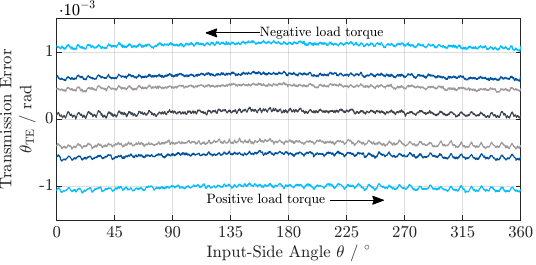"}
		\caption{Position Domain}
	\end{subfigure}
	\hfill
	\begin{subfigure}{0.495\textwidth}
		\includegraphics[scale=1]{"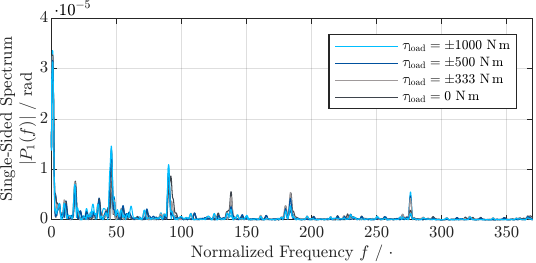"}
		\caption{Spatial Frequency Domain}
	\end{subfigure}
	\caption{Comparison of the transmission errors at different load torques for \(\dot{\theta}=\qty{-2.5}{\degree\per\second},~T\approx\qty{27}{\degreeCelsius}\).}
	\label{fig:meas_torque}
\end{figure*}

The characteristics of the TEs are analyzed based on an identification data set collected using the described experimental setup~(\autoref{ssec:experimental_setup}).
In total~\num{224} measurements~\cite{bauerReplicationDataModeling2025man} with all combinations of the properties
\begin{itemize}
\item \(\tau_\mathrm{load} \in \left\{\qty{0}{\N\m}, \pm\qty{333}{\N\m}, \pm\qty{500}{\N\m}, \pm\qty{1000}{\N\m}\right\}\),
\item \(\dot{\theta} \in \left\{\pm\qty{2.5}{\degree\per\second}, \pm\qty{5.0}{\degree\per\second}, \pm\qty{7.5}{\degree\per\second}, \pm\qty{10.0}{\degree\per\second} \right\}\), and
\item \(T \in \left\{\qty{20}{\degreeCelsius}, \qty{30}{\degreeCelsius}, \qty{40}{\degreeCelsius}, \qty{50}{\degreeCelsius} \right\} \)
\end{itemize}
were conducted.
Since the gearbox cannot be cooled in the current setup, the measurements with a desired temperature of~\qty{20}{\degreeCelsius} were carried out at the current lubricant temperature upon starting the measurement program.
The influence of torque ripple from the load-side motor was filtered out of the measurement data, as it would not be present in a robot joint.
This was accomplished by applying a notch filter to the harmonic frequencies~\(\num{6}\cdot 25 = 150\) and~\(\num{12}\cdot 25 = 300\) associated with the~\num{25} pole pairs in the load-side motor.\\
In \autoref{fig:meas_torque}, the influence of the load torque~\(\tau_{\mathrm{load}}=\{\qty{-1000}{\N\m},\dots,\qty{1000}{\N\m}\}\) for measurements with a velocity of~\(\dot{\theta}=\qty{-2.5}{\degree\per\s}\) and a temperature of approximately~\(T\approx\qty{27}{\degreeCelsius}\) is shown.\begin{figure*}
	\centering
	\begin{subfigure}{0.495\textwidth}
		\includegraphics[scale=1]{"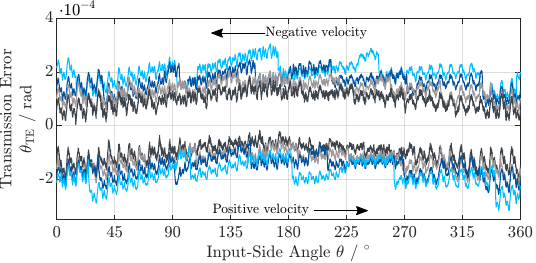"}
		\caption{Position Domain}
	\end{subfigure}
	\hfill
	\begin{subfigure}{0.495\textwidth}
		\includegraphics[scale=1]{"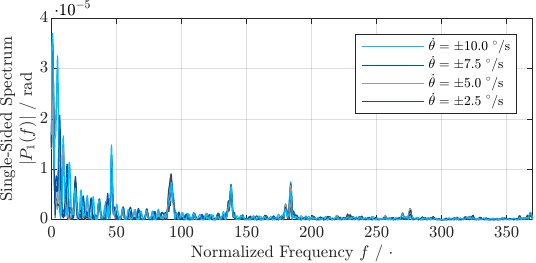"}
		\caption{Spatial Frequency Domain}
	\end{subfigure}
	\caption{Comparison of the transmission errors at different velocities for \(\tau_\mathrm{load} = \qty{0}{\N\m},~T\approx\qty{26}{\degreeCelsius},\dots,\qty{28}{\degreeCelsius}\).}
	\label{fig:meas_velocity}
\end{figure*}\begin{figure*}
	\centering
	\begin{subfigure}{0.495\textwidth}
		\includegraphics[scale=1]{"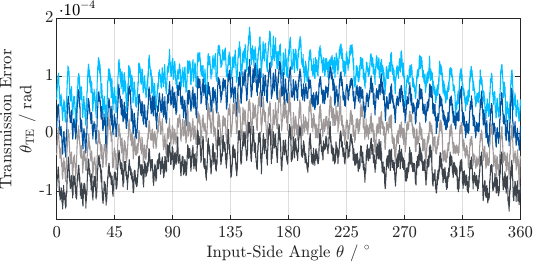"}
		\caption{Position Domain}
	\end{subfigure}
	\hfill
	\begin{subfigure}{0.495\textwidth}
		\includegraphics[scale=1]{"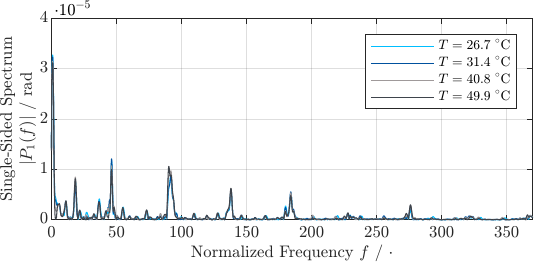"}
		\caption{Spatial Frequency Domain}
	\end{subfigure}
	\caption{Comparison of the transmission errors at different temperatures for \(\tau_\mathrm{load} = \qty{0}{\N\m},~\dot{\theta}=\qty{-2.5}{\degree\per\second}\).}
	\label{fig:meas_temp}
\end{figure*}
It can be seen that the load torque has a large influence on the constant offset, which can be explained by the compliance of the gear.
In addition, changes in the amplitudes and phases of the TEs can be observed.
For velocities~\(\dot{\theta}=\{\qty{-10}{\degree\per\second},\dots,\qty{10}{\degree\per\second}\}\) at~\(\tau_{\mathrm{load}}=\qty{0}{\N\m}\) load torque and temperatures in between~\(T\approx\qty{26}{\degreeCelsius}\) and~\(T\approx\qty{28}{\degreeCelsius}\) these changes in amplitude and phase can also be seen, while the influence on the constant offset is less distinct, but still present~(\autoref{fig:meas_velocity}).
In~\autoref{fig:meas_temp} the influence of the temperature is visualized by comparing the measurements at a temperature of~\(T\approx\{\qty{27}{\degreeCelsius},\dots,\qty{50}{\degreeCelsius}\}\) at load torque \(\tau_\mathrm{load} = \qty{0}{\N\m}\) and velocity \(\dot{\theta}=\qty{-2.5}{\degree\per\second}\).
Here, only minor temperature influences on the amplitudes are noticeable, while changes in the constant offset and the phases are more relevant.\\
Analyzing the single-sided spectrum of the TE measurements and comparing the dominant peaks to the mechanical properties of the CD described in~\autoref{ssec:experimental_setup}, the following mechanical frequencies were selected for the identification of the Fourier series parameter:
\begin{itemize}
	\item Output shaft rotation: \[f=i,~i\in\{1, 3, 6, 9, 13, 14, 18, 19, 27\}\mathrm{.}\]
	Most of the low frequency errors are assumed to be related to angular errors of the output shaft and occur at multiples of the period.
	\item Planetary gear stage: \[f=i\cdot u_1,~i\in\{1, 3\}\mathrm{.}\]
	Two harmonic frequencies with noticeable amplitudes can be related to the reduction ratio of the planetary stage and therefore it is concluded that there are caused by manufacturing and assembly tolerances in this stage.
	\item CD stage: \[f=i\cdot u_2,~i\in\{1, 2, 3, 4, 6\}\mathrm{.}\]
	For these six harmonic frequencies of the CD stage, it is not possible to distinguish whether they are related to manufacturing tolerances of the~\(z_4=46\) pins or assembly tolerances between the cycloidal disc and the pins.
	\item Input gear: \[f=i\cdot z_1,~i\in\{1\}\mathrm{.}\]
	One harmonic frequency is related to tolerances of the input gear.
	\item Cycloidal disc: \[f=i\cdot z_3,~i\in\{2, 3, 4\}\mathrm{.}\]
	These harmonic frequencies are possibly related to deformations and tolerances in the cycloidal disc.
\end{itemize}
The polynomial Fourier series model of the TEs is identified by applying the method described in~\autoref{ssec:identification} and using these selected frequencies.
An analysis of the parameter sensitivities, defined as the parameter change required to double the model error, showed an almost similar sensitivity of the parameters~\(p_{\cdot,k,i}\) for the linear terms~\(i=\{4,\dots,6\}\), the two-factor coupling terms~\(i=\{7,\dots,9\}\), and the constant term~\(i=\{10\}\). The quadratic term parameters~\(i=\{1,\dots,3\}\) were more sensitive to parameter changes due to the higher order of these terms. They were most sensitive to parameter changes in the torque coefficient~\(i=\{1\}\) and least sensitive to changes in the temperature coefficient~\(i=\{3\}\).
The root mean square error~(RMSE) and the mean absolute error~(MAE) are used as measures of the errors between measurement and polynomial Fourier series model.
To provide an overall error indicator, the mean RMSE and the mean MAE are calculated per direction.
The mean of the measurement and the constant offset~\(a_0\) are not included here because they would distort the comparability between the different measurements.
In positive direction, the mean RMSE is~\qty{2.76e-5}{\radian} and the mean MAE is~\qty{2.23e-5}{\radian}, while in negative direction, the mean RMSE is~\qty{3.01e-5}{\radian} and the mean MAE is~\qty{2.44e-5}{\radian}.

\subsection{Model Validation}
\begin{figure*}
	\centering
	\begin{subfigure}{0.495\textwidth}
		\includegraphics[scale=1]{"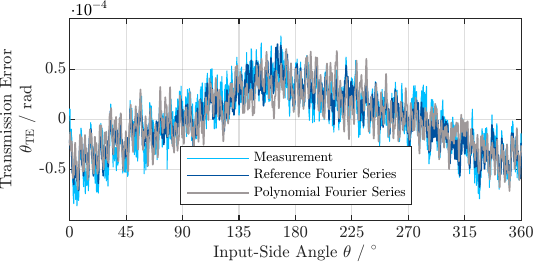"}
		\caption{Position Domain}
	\end{subfigure}
	\hfill
	\begin{subfigure}{0.495\textwidth}
		\includegraphics[scale=1]{"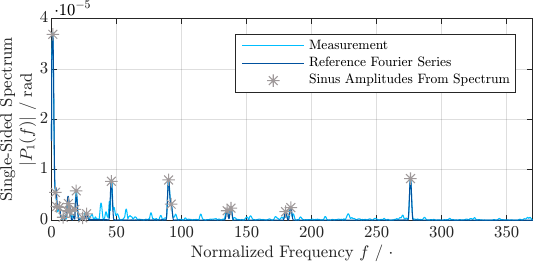"}
		\caption{Spatial Frequency Domain}
	\end{subfigure}
	\caption{Exemplary model validation for \(\tau_\mathrm{load} = \qty{-750}{\N\m},~\dot{\theta}=\qty{2.5}{\degree\per\second},~T=\qty{26}{\degreeCelsius}\).}
	\label{fig:validation}
\end{figure*}

To validate the identified model, \(19\) validation measurements~\cite{bauerReplicationDataModeling2025man} were conducted. The measurement parameters~(load torque~\(\tau_{\mathrm{load}}\), velocity~\(\dot{\theta}\), temperature~\(T\)) were selected using Sobol's quasi-random sequence generator~\cite{joeRemarkAlgorithm6592003}, given the bounds of the minimum and maximum values used in the identification process.
This approach guarantees that the validation data set is independent of the identification data set.
The suitability of the chosen model approach for extrapolation has not been evaluated.
In \autoref{fig:validation}, the results for an example validation measurement with load torque~\(\tau_\mathrm{load} = \qty{-750}{\N\m}\), velocity~\(\dot{\theta}=\qty{2.5}{\degree\per\second}\) and temperature~\(T=\qty{26}{\degreeCelsius}\) are visualized.
A reference Fourier series was computed from this single measurement and represents the theoretical optimum that the polynomial Fourier series model computed from the measurement data set could achieve.
In the single-sided amplitude spectrum~(b), the amplitudes at the selected harmonic frequencies are marked.
It can be seen that these frequencies are included in the reference Fourier series calculation, while other frequency content of the signal is neglected.
This is evident in the position domain representation~(a), where the reference Fourier series fits the general behavior, while some peaks are not included.
For the example measurement, a RMSE of~\qty{1.83e-5}{\radian} and a MAE of~\qty{1.49e-5}{\radian} results.
Comparing the reference Fourier series with the output of the polynomial Fourier series model reveals some discrepancies.
Due to the simplification given by a quadratic polynomial, the~\(A_k (\tau_\mathrm{load}, \dot{\theta}, T)\) and~\(\varphi_k (\tau_\mathrm{load}, \dot{\theta}, T)\) derived from the polynomial cannot perfectly fit the~\(A_k\) and~\(\varphi_k\) from a single measurement.
For the validation measurements in positive direction, the mean RMSE is~\qty{2.40e-5}{\radian} and the mean MAE is~\qty{1.92e-5}{\radian}.
In negative direction, the mean RMSE is~\qty{2.82e-5}{\radian} and the mean \hbox{MAE is~\qty{2.26e-5}{\radian}}.

\section{Discussion}
In this contribution it has been shown that the mechanical properties of CDs can be related to the relevant harmonic frequencies of the TEs.
Only these harmonic frequencies have been considered in the Fourier series calculation.
Fourier series have been shown to be a valid model structure for the mainly periodic TEs.
It takes into account load torque, velocity, and temperature, which has not been found for polynomial Fourier series models in the state of the art.
It is possible, that the velocity-dependent changes in the TEs are caused by changes in the velocity-dependent friction torque, and are therefore also a form of load-dependent effect.\\
The choice of a Fourier series as a model structure neglects any non-periodic TE components that may be present.
Such non-periodic errors can occur if individual pins or teeth in the CD are damaged. Therefore, even a Fourier series including all possible frequencies (not just these derived from mechanical properties) could not achieve a perfect fit.
It would be possible to include higher harmonics than those selected to improve the fit, but they would contribute little to the overall result.
Another option would be to use a higher order polynomial that includes higher order coupling terms between load torque, velocity, and temperature inputs.
In general, it is desirable to keep the number of model parameters and model complexity as low as possible to avoid overfitting and to ensure real-time computation.
A sensitivity analysis for higher order polynomials should be performed in further research to identify the most relevant parameters.
As an alternative to the presented method, it would be possible to model the TEs with a machine learning approach, as has been shown for rack-and-pinon drives~\cite{steinleLearningCompensationStateDependent2024}.
This alternative approach could provide a better TE fit due to its ability to include non-periodic effects.\\
The presented model requires only the input-side angle from the motor encoder and does not depend on the load-side angle after identification, which is an improvement over the presented state of the art~(\autoref{ssec:StateOfTheArt}).
As presented, the approach still relies on information from the load-side torque sensor and the lubricant temperature sensor in the CD.

\section{Outlook}
\begin{figure}[b]
	\centering
	\input{"images/block_diagram.tex"}
	\caption{Compensation approach in the system model.}
	\label{fig:BlockDiagramOverview}
\end{figure}
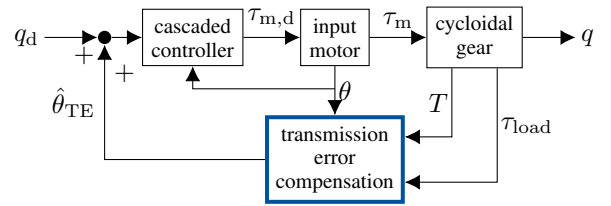

The authors propose to use the presented TE model to compensate the resulting position error of the load-side angle with a feedforward control approach as shown in~\autoref{fig:BlockDiagramOverview}.
Since the overall goal is to increase the path accuracy of the entire robot, the goal at the axis level is to increase the accuracy of the load-side angle~\(q\).
Thus, the model output~\(\hat{\theta}_\mathrm{TE}\) is added to the desired load-side angle~\(q_\mathrm{d}\).
The correction term is propagated through the cascaded controller, resulting in a set torque~\(\tau_\mathrm{m,d}\) for the input-side motor that implicitly includes the correction.
The resulting torque~\(\tau_\mathrm{m}\) acts on the input side of the CD and a corrected output-side angle follows.
In the suggested compensation structure, the velocity~\(\dot{\theta}\) is derived from the input-side motor angle~\(\theta\).
The load torque~\(\tau_\mathrm{load}\) and the temperature~\(T\) are measured directly.
In quasi-static operation, it can be assumed that~\(\tau_\mathrm{load}\approx\tau_\mathrm{m}\) and the~\(\tau_\mathrm{m}\) derived from the motor current measurement can be used as an alternative input for the TE model.
In a further development step, the measurements for the load-side torque and the lubricant temperature sensor could be replaced by observer structures.
For the load-side torque sensor of the experimental setup, this was presented in~\cite{mesmerInvestigationCompensationHysteresis2023}.
In further research, the model needs to be evaluated in closed-loop operation on the experimental setup.\\
While the presented polynomial Fourier series model represents a partial solution to the problem of increasing path accuracy in industrial robots, the larger influence of compliance~\cite{rendersKinematicCalibrationGeometrical1991} must also be considered. Therefore, a combination with the compliance compensation approach for CDs presented in~\cite{mesmerInvestigationCompensationHysteresis2023} is intended.

\section*{Data Availability}
The measurement data used to identify the polynomial Fourier series is available on DaRUS,  the data repository of the University of Stuttgart: \url{https://doi.org/10.18419/DARUS-4454}

\bibliographystyle{IEEEtran}
\bibliography{IEEEabrv,ICIT_2025_BibTeX,ICIT_2025_manual}

\end{document}

%% file: images/block_diagram.tex
\begin{tikzpicture}
	\node[punkt] (sum1) {};
	\node[block, right=0.4cm of sum1] (controller) {cascaded\\controller};
	\node[block, right=0.75cm of controller] (motor) {input\\motor};
	\node[block, right=0.75cm of motor] (gear) {cycloidal\\gear};
	
	\node[blueblock, below=0.65cm of motor] (compensation) {transmission\\error\\compensation};
	
	\node[input, left=0.7cm of sum1] (input1) {};
	\node[input, right=0.7cm of gear] (output1) {};
	
	\draw[->] (input1) -- node[near end, below] {\(+\)} node[left, xshift=-0.3cm] {\(q_\mathrm{d}\)} (sum1);
	\draw[->] (sum1) -- (controller);
	\draw[->] (controller) -- node[midway, above] {\(\tau_{\mathrm{m,d}}\)} (motor);
	\draw[->] (motor) -- node[midway, above] {\(\tau_{\mathrm{m}}\)} (gear);
	\draw[->] (gear) -- node[right, xshift=0.3cm] {\(q\)} (output1);
	\draw[->] (motor) -- node[midway, right, xshift=-0.075cm] {\(\theta\)} (compensation);
	\draw[->] ($(gear.south)+(-0.3,0)$) |- node[near start, left, xshift=0.075cm] {\(T\)} ($(compensation.east)+(0,0.3)$);
	\draw[->] ($(gear.south)+(+0.3,0)$) |- node[near start, right, xshift=-0.075cm] {\(\tau_{\mathrm{load}}\)} ($(compensation.east)+(0,-0.3)$);
	\draw[->] (compensation) -| node[near end, left] {\(\hat{\theta}_\mathrm{TE}\)} node[very near end, right] {\(+\)} (sum1);
	
	\draw[->] ($(motor.south)+(0,-0.3)$) -| (controller);
\end{tikzpicture}